\documentclass[11pt,a4paper]{article}
\usepackage[hyperref]{emnlp-ijcnlp-2019}
\usepackage{times}
\usepackage{latexsym}
\usepackage{url}

\usepackage{graphicx}
\usepackage{amsmath}
\usepackage{bm}
\usepackage[group-separator={,},group-minimum-digits={3}]{siunitx}
\usepackage{subfigure}
\usepackage{amssymb}
\usepackage{multirow}
\usepackage{xcolor}

\usepackage{enumitem}

\setitemize[1]{itemsep=5pt,partopsep=5pt,parsep=\parskip,topsep=5pt}

\title{Group, Extract and Aggregate: Summarizing a Large Amount of \\ Finance News for Forex Movement Prediction}

\author{Deli Chen\textsuperscript{1}\thanks{~This work is done when Deli Chen is a intern at Mizuho Securities.},
Shuming Ma\textsuperscript{1},
Keiko Harimoto\textsuperscript{2},
Ruihan Bao\textsuperscript{2},
\textbf{Qi Su\textsuperscript{1},
Xu Sun\textsuperscript{1}}
\\ \\
\textsuperscript{1}{MOE Key Lab of Computational Linguistics, School of EECS, Peking University}\\
\textsuperscript{2}{Mizuho Securities Co., Ltd.}\\
\{chendeli,shumingma,sukia,xusun\}@pku.edu.cn,\\
\{keiko.harimoto,ruihan.bao\}@mizuho-sc.com 
}

\begin{document}
\maketitle

\begin{abstract}
Incorporating related text information has proven successful in stock market prediction. However, it is a huge challenge to utilize texts in the enormous forex (foreign currency exchange) market because the associated texts are too redundant. In this work, we propose a BERT-based Hierarchical Aggregation Model to summarize a large amount of finance news to predict forex movement. We firstly group news from different aspects: time, topic and category. Then we extract the most crucial news in each group by the SOTA extractive summarization method. Finally, we conduct interaction between the news and the trade data with attention to predict the forex movement. The experimental results show that the category based method performs best among three grouping methods and outperforms all the baselines. Besides, we study the influence of essential news attributes (category and region) by statistical analysis and summarize the influence patterns for different currency pairs.


\end{abstract}

\section{Introduction}
\begin{figure}[t]
\centering
\includegraphics[scale=0.3]{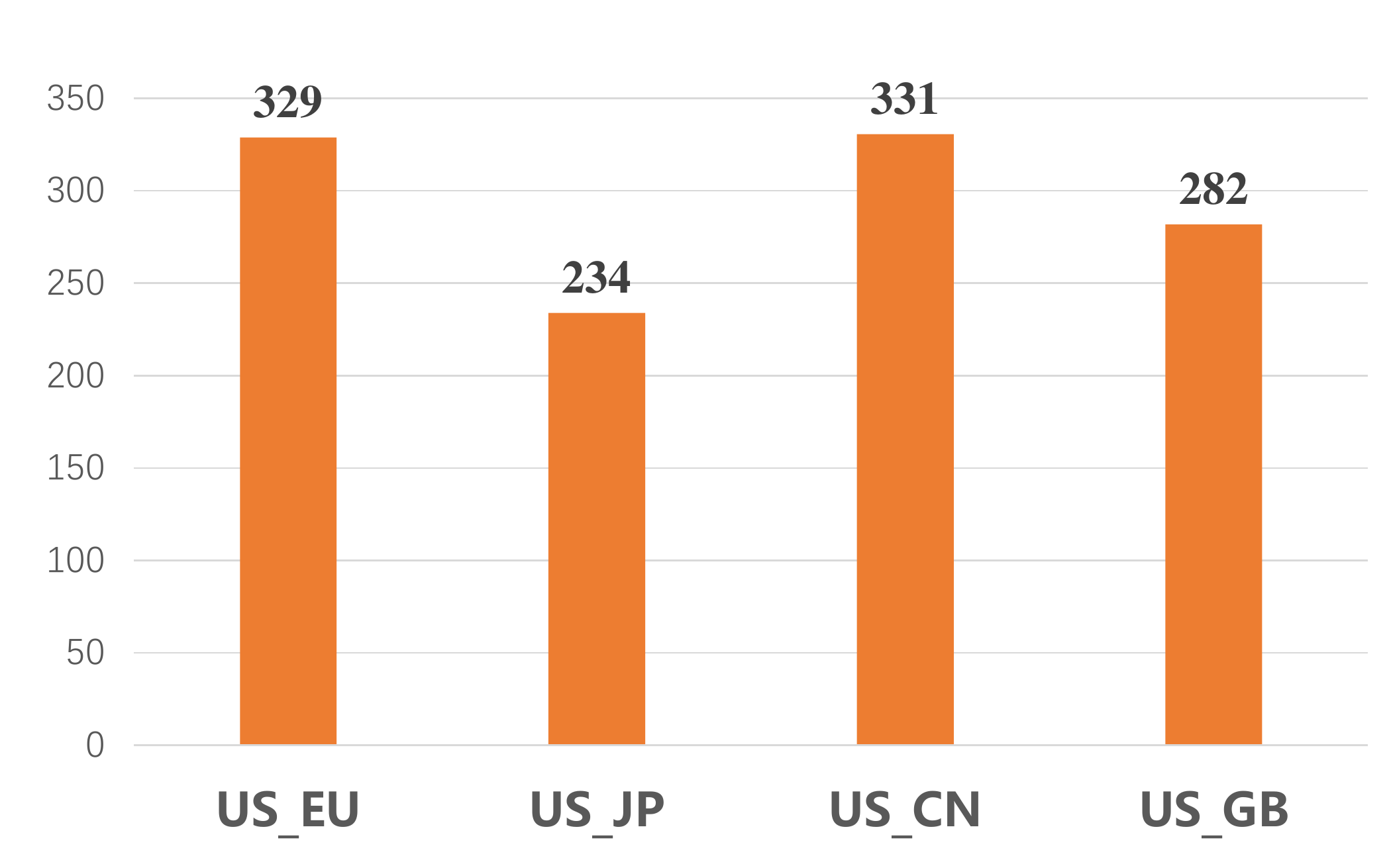}
\vspace{-0.1in}
\caption{Average numbers per hour of forex related news from Reuters in 2013-2017. $\rm US\_EU$ represents news related to US, Europe or both of them.}
\label{news_count}
\vspace{-0.25in}
\end{figure}

Deep learning and Natural Language Processing technologies have been widely applied in market prediction tasks~\citep{StockOther2Intra,stock_other_breaknews,Raw3MultiModal2AAAI15TensorMethod,dl_crnn}, and the market related finance news has proven very useful for the prediction~\citep{stock_ding_event3,stock_acl19_vae}. However, the studies of prediction in forex market, which is the largest market in the world with the highest daily trading volume, is much less than that in the stock market. 
Figure~\ref{news_count} shows the average numbers per hour of forex related news. There is a large amount of finance news related to forex trading with different influence, so it is a huge challenge to extract the useful semantic information from news.
Most of previous works~\citep{ml_direction_change,ml_autoencoder_svr,ml_random_forest,ml_elastix_network,ml_black_region} on forex prediction ignore related text totally and focus on the forex trade data only, which loses the important semantic information. Yet existing works~\citep{text_wsd_sa,text_dimension_reduce} applying finance news in forex prediction mainly rely on manual rules to build feature vectors, which can hardly access the semantic information effectively. 

To make better use of finance news, we propose a novel neural model: Bert-based Hierarchical Aggregation Model (\textbf{BHAM}) to summarize a large amount of finance news for forex movement prediction. 
We suppose that the finance news is redundant and only a small amount of news plays a crucial role in forex trading. 
So the key point is how to extract the most important news. 
In BHAM, we design a hierarchical structure to extract essential news at the group level first and then aggregate the semantic information across all groups. We expect the news is more related intra-group and less related inter-groups to make the extraction more effective.
We design three grouping methods from different aspects: time, topic or category.  
At the group level, we concatenate news headlines in the same group and regard news extraction in each group as an extractive summarization task. We modify the SOTA extractive summarization model proposed in~\citep{summary_bert_select} to select the most important news. The connection process can let the selected news both content aware and context aware.
Followingly, we conduct multi-modal interaction between news data and trade data through attention mechanism to predict the forex prediction.
The trade data represents the history movement of the forex, and the news data represents the environment variable. These two types of information are highly related. 

We conduct experiments on four major currency pairs (USD-EUR, USD-JPY, USD-RMB, USD-GBP), and the experimental results show that the category-based BHAM performs best among all the baselines and proposed methods in all currency pairs. Based on this method, we analyze the influence of input time and prediction time on forex trading. We also analyze the influence of news category and news region and find various influence patterns for different currency pairs, which may be enlightening to the forex investors. The main contributions of this works are summarized as follows:
\begin{itemize}
    \item We design a novel neural model to incorporate finance news in forex movement prediction. To the best of our knowledge, this is the first work to use the neural model to summarize a large amount of news for forex movement prediction.
    \item We propose three news grouping methods from different aspects: time, topic and category. Experiments show that the category based method performs best and outperforms all the baselines.
    \item Based on our experiments, we study the effect of time parameters on forex trading. We also analyze and summarize different influence patterns of finance news (both category and region) on different currency pairs.  
\end{itemize}

\section{Related Work}
BERT~\citep{bert} is a potent pre-trained contextualized sentence representation and has proven obvious improvement for many NLP tasks~\citep{bert_app1,bert_app2}.~\citet{summary_bert_select} proposes a modified BERT for extractive summarization and achieve the state-of-the-art result in extractive document summarization task. 

There have been many studies applying the related text in market prediction tasks. Moreover, the text assisted stock movement prediction has attracted many researchers' interest. Most of these works predict stock movement based on single news:~\citet{Raw2Emnlp14sentiment} utilize the sentiment analysis to help the prediction. \citet{Raw1Coling18NewsBody} adopt the summarization of news body instead of headline to predict. \citet{stock_ding_event3} propose the knowledge-driven event embedding method to make the forecast. Yet some others choose multi-news:~\citet{stock_hann} propose a hybrid attention network to combine news in different days.
However, the number of combined news is still limited and much smaller than that of forex news.

Compared to stock prediction, works about forex prediction is much scarce, and most of these works~\citep{rl_fx,ml_direction_change,dl_neuro_fuzzy,ml_system_genibux,ml_elastix_network,ml_black_region} do not consider the text information.
~\citet{ml_autoencoder_svr} employ stacked autoencoder to get the trade data representation and adopt support vector regression to predict.~\citet{ml_svm_ga} combine SVM with genetic algorithms to optimize investments in Forex markets based on history price.~\citet{dl_cnn} choose the convolutional neural network to process the trading data.
Besides, only limited works utilize the forex related text in the prediction process.~\citet{text_dimension_reduce} adopt the WordNet~\citep{WordNet} and SentiWordNet~\citep{sentiWordNet} to extract the text semantic and sentiment information and build the text feature vector to forecast forex movement. Following this work,~\citet{text_wsd_sa} add word sense disambiguation in the sentiment analysis of news headlines.~\citet{text_j48} apply the J48 algorithm in analyzing text. This kind of method pays more attention to access a fixed feature vector from news and can only represent news on a shallow level. 
In this work, we propose a selection and aggregation neural framework to process the larger amount of finance news and employ the powerful pre-trained BERT as text encoder, which can learn the deep semantic information effectively.

\begin{figure*}[t]
\centering
\includegraphics[scale=0.42]{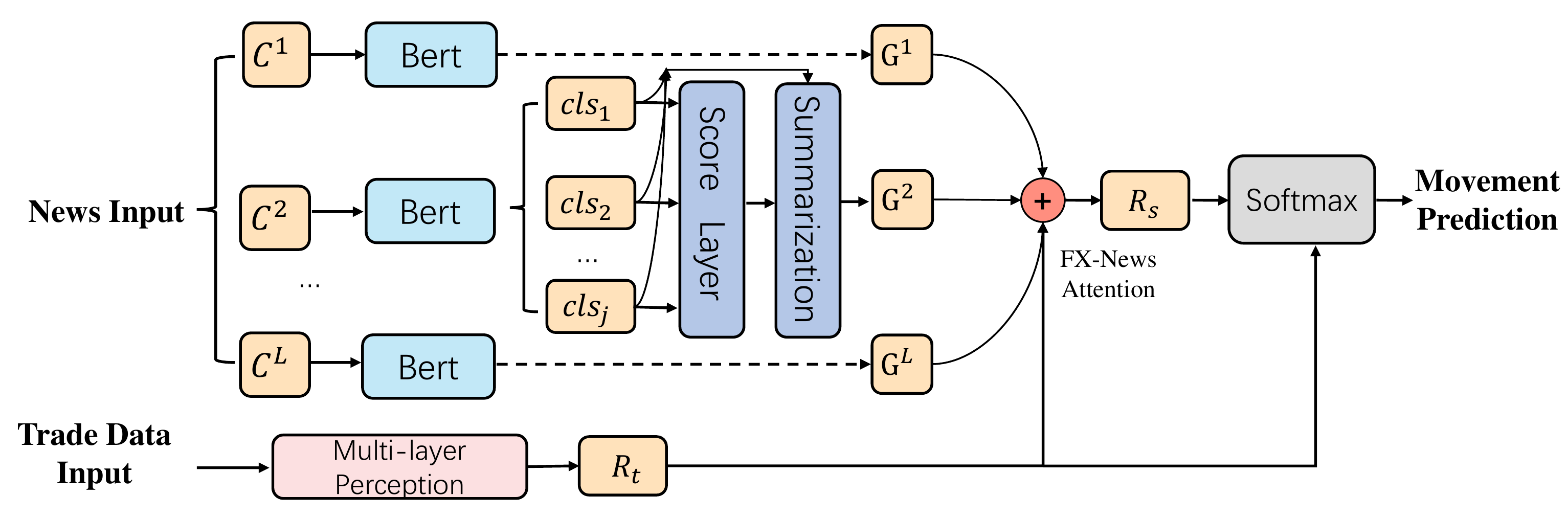}
\vspace{-0.1in}
\caption{The overview of the proposed model.}
\label{model_overview}
\vspace{-0.1in}
\end{figure*}

\begin{figure*}[t]
\centering
\includegraphics[scale=0.42]{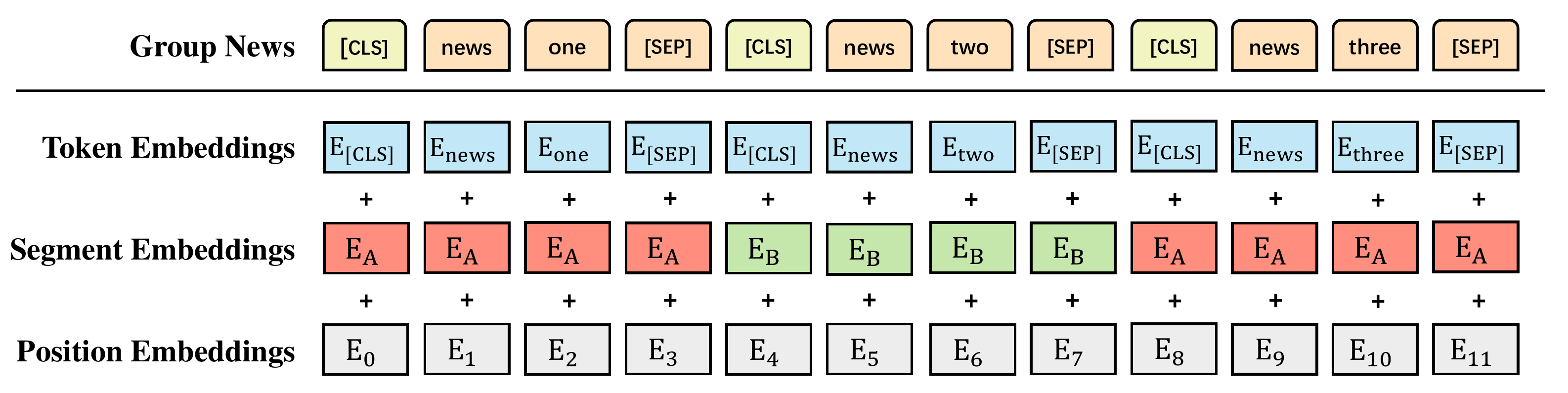}
\vspace{-0.1in}
\caption{The BERT input in each news group.}
\label{bert_input}
\vspace{-0.2in}
\end{figure*}

\section{Approach}
\subsection{Problem Formulation}
Each sample in the dataset $(x,y,f)$ contains the set of news text $x$, the forex trade data $y$, and the forex movement label $f$. $x$ and $y$ happen in the same input time window. To be more specific, $x$ is a list of news groups $x=\left \{ C^1,C^2,\cdots,C^L  \right \}$. $L$ is the number of groups. The methods for dividing groups are introduced in Section~\ref{section_news_groups}. Each news group is a sequence of finance news $[news_1,news_2,\cdots,news_K]$ in chronological order. $y$ is the trade data embedding accessed by the method introduced in Section~\ref{section_trade_data_embedding}. And $f \in \left \{ 1, 0 \right \}$ is the forex movement label telling whether the forex trade price is up or down after a certain time (we call it prediction delay). The forex movement prediction task can be defined as assigning movement label for the news input and trade data input.

\subsection{Model Overview}
The overview of the Bert-based Hierarchical Aggregation Model (\textbf{BHAM}) is displayed in Figure~\ref{model_overview}. The model can be generally divided into two steps: (1)Intra-group extraction and (2)Inter-groups aggregation. In the Intra-group extraction step, news in the same group is connected as a continuous paragraph, and we conduct extractive summarization on this paragraph to select the most important news. Specifically, we employ BERT as the encoder to get the contextualized paragraph representation and compute the importance score for each news. Then we select and aggregate the top-k (k is a hyper-parameters) news to get the final group representation. In the Inter-groups aggregation step, we first access the trade data representation by a 3-layer perceptron and then employ the trade data representation as a query to calculate the attention scores of all the news group and obtain the final news representation. Finally, we fuse the final news representation and the trade data representation to predict the forex movement. 

\subsection{Intra-group Extraction}
There will be lots of news in the same group, and we suppose that only a small amount of news has the greatest influence on the forex movement. The purpose of this step is to select the essential news from all news in group, which is redundant and full of noise. Inspired by the BERT-based extractive summarization model proposed in~\citep{summary_bert_select}, we modify this method to select the most crucial news in each group. All the news in the same group is related to the subject of this group, and the connection of them in chronological order can be regarded as the continuous description of the group subject. The connection can make the news representations realize the context information of this group by passing information among different news. We suppose the context information can help select better news in group.

The form of group news input for BERT encoder is illustrated in Figure~\ref{bert_input}. We insert a $[$CLS$]$ token before each news and a $[$SEP$]$ token after each news. For the segment embedding, we use the loop of $[E_A, E_B]$ to extend the raw segment embedding of BERT to multi-sentences. After the BERT encoding, all the $[$CLS$]$ tokens $\bm{cls}$ are regarded as the semantic representations of the corresponding news. The importance score for each news is calculated base on these $[$CLS$]$ tokens:
{\setlength{\abovedisplayskip}{5pt}
\setlength{\belowdisplayskip}{5pt}
\begin{flalign}
 \bm{score^i} &= {\rm sigmoid}(\bm{W_0}*\bm{cls^i}+\bm{b_0})  \\
 \bm{t^i} &= {\rm TOP_k}(\bm{score^i}) \\
 \bm{s}^i &= {\rm softmax}(\bm{t^i}) 
\end{flalign}}
Where $i\in\{1,2,\cdots,L\}$, $L$ is the number of groups. $\bm{cls^i}$ is the list of $[$CLS$]$ tokens in the i-th group. $\bm{W_0}$ and $\bm{b_0}$ are the trainable parameters. $\bm{score^i}$ is a list of values indicating the important scores of news. ${\rm TOP_k}$ is an operation to select the top-k pieces of news with the highest scores. Then the group representation is calculated by the weighted sum of the top-k $[$CLS$]$ tokens:
{\setlength{\abovedisplayskip}{5pt}
\setlength{\belowdisplayskip}{5pt}
\begin{flalign}
\bm{G^i} = \sum_{j=1}^{k} \bm{cls}^i_{j}* s^i_{j}   \label{intra-weight}
\end{flalign}}
The $\bm{G^i}$ is the final representation of the i-th news group which contains the semantic information from the most important news in this group.

\subsection{Inter-groups Aggregation}
The purpose of this step is to aggregate semantic information at the inter-groups level. The forex trade data and the finance news are highly relevant: the trade data represents the history movement of forex, and the finance news represents the environmental variable. So the combination of them can help us model the forex movement better. In a certain input time, news groups have different impacts on forex movement. So we employ the trade data as a query to calculate the attention weights of news groups. Then the weighted sum of news groups and the trade data representation are finally fused to predict the forex movement.
For forex trade data $y$, we apply a 3-layer perceptron to access the trade data representation $\bm{R_t}$, and each layer is a non-linear transform with $\rm Relu$ activation function.
Then we calculate the attention weight between $\bm{R_t}$ and $\bm{G^i}$ :
{\setlength{\abovedisplayskip}{5pt}
\setlength{\belowdisplayskip}{5pt}
\begin{flalign}
& g(i) ={\rm Relu}(\bm{R_t}* \bm{W_a} * \bm{G^{i\top}})  \\
& att_i = \frac{e^{g(i)}}{\sum_{i=1}^{L}e^{g(i)}} \label{co-attention}
\end{flalign}}
Where $att(i)$ is the i-th news group's attention weight to trade data. Then we sum the news groups representations up to get the final news semantic representation $\bm{R_s}$:
{\setlength{\abovedisplayskip}{5pt}
\setlength{\belowdisplayskip}{5pt}
\begin{flalign}
&\bm{R_s} = \sum_{i=1}^{L} \bm{G^i}*att_i 
\end{flalign}}
To fuse the news semantic and trade data representations effectively, we choose the fusion function used in~\citep{Attention1Main,FusionFunction} to fuse $\bm{R_s}$ and $\bm{R_t}$ and predict the movement:
{\setlength{\abovedisplayskip}{5pt}
\setlength{\belowdisplayskip}{5pt}
\begin{flalign}
&\bm{R} = [\bm{R_t};\bm{R_s};\bm{R_t}-\bm{R_s};\bm{R_t}\circ \bm{R_s}]  \\
& \widehat{p}(f|x,y) = {\rm softmax}(\bm{W_p} * \bm{R} + \bm{b_p})
\end{flalign}}
$\circ$ means element-wise multiplication. 

\subsection{Methods of Grouping News}
\label{section_news_groups}
In this part, we introduce the three news grouping methods. The ideal division enables news groups to be high cohesion and low coupling, which means the semantic information of finance news should be highly related intra-group and less related inter-groups. We suppose that extracting news by groups can reduce the extraction difficulty compared to extracting from all news directly because news in the same group is close to each other and has less noise. Moreover, this method can help us analyze the contributions of different groups.

\subsubsection{Grouping by Time}
In this method, finance news is divided into groups according to the time when news happens. We set the time unit to 5 minutes and news released in the same time unit will be divided into the same group. This method supposes that news happened closely is highly correlated.

\subsubsection{Grouping by Topic}
In this method, finance news is divided into groups by news topic. The news topics are generated by unsupervised news clustering. In this work, we choose the affinity propagation algorithm~\citep{AP_cluster} to generate news clusters without setting the number of clusters subjectively. Moreover, we choose the tf-idf of 2-gram features from news headlines.
This method supposes that finance news focuses on several finance event topics at a particular time. News in the same topic describes this topic from different aspects and has a high correlation.

\subsubsection{Grouping by Category}
In this method, news is divided into groups by category. The news categories\footnote{Use the Reuters professional financial news category(\url{https://liaison.reuters.com/tools/topic-codes}) and merge some similar categories.} are $\{$Business Sectors, Business General, Business Assets, Business Commodities, Business Organizations, Politics\&International Affairs, Arts\&Culture\&Entertainment\&Sports,  Science \&Technology, Other$\}$. This method supposes that news in the same category is close to each other.

\subsection{Trade Data Embedding}
\label{section_trade_data_embedding}
The raw record of forex data includes the open/ close/high/low trade prices for each minute. In order to extract all the possible features, we build the trade data embedding $y$ containing multi aspects:
\begin{itemize}
    \item \textbf{Raw Number}: open/close/high/low trade price for each trade minute.
    \item \textbf{Change Rate}: change rate of open/close/ high/low price compared to last trade minute.
    \item \textbf{Trade Statistics}: mean value, max value, min value, median, variance of all the trade prices in input minutes.
\end{itemize}
The min-max scale is applied for each currency pair's samples to scale the raw numbers in $y$ to $[0,1]$ according to the maximum and minimum value of each feature.

\subsection{Training Objective}
The loss function of the proposed model includes two parts: the negative log-likelihood training loss and the $L_2$ regularization item:
{\setlength{\abovedisplayskip}{5pt}
\setlength{\belowdisplayskip}{5pt}
\begin{flalign}
&Loss = - f * log\,p(f|x,y,\theta) +  \frac{\lambda }{2} \left \| \theta    \right \|^2_2 
\end{flalign}}
$\theta$ is the model parameters. 
Experiments show that the performance improves after adding $L_2$ regularization.
We train three models with different news grouping methods: time, topic and category, and we call them BHAM-Time, BHAM-Topic, BHAM-Category, respectively.

\section{Experiment}
\subsection{Dataset}
The experiment dataset is accessed from the professional finance news providers Reuters\footnote{Source Reuters News cThomson Reuters cREFINITIV, \url{https://www.thomsonreuters.com/en.html}}. We collect forex trade data of four major currency pairs (USD-EUR, USD-JPY, USD-RMB, USD-GBP) from 2013 to 2017. We collect the open/close/high/low trade price for each trade minute. As for the finance news data, we collect all the English news happened in trade time released by Reuters and match the news with target currency pairs according to news region. For example, we match USD-EUR with news related to US, Europe or both of them. The raw data contains both news headline and body, and we utilize the headline only since the headline contains the most valuable information and has less noise. The forex movement label $f$ is decided by the comparison of prediction time price and the input window ending price. 
We design the symbol USD-EUR(20-10) to represent the prediction for the USD-EUR exchange rate with 20 minutes input time and 10 minutes prediction delay.
To access more data for training, we overlap the input time of samples. For example, when overlap-rate is 50$\%$, two consecutive samples' input time will be 8:00-8:20 am and 8:10-8:30 am. Then the data samples will be twice as large as no overlap condition (In the USD-EUR(20-10) dataset, the number of samples will increase from 31k to 62k). We reserve 5k samples for developing and 5k samples for testing. All the rest of samples are applied for training.   

\subsection{Experiment Setting}
We choose the pytorch-pretrained-BERT\footnote{\url{https://github.com/huggingface/pytorch-pretrained-BERT}} as BERT implement and choose the bert-base-uncased version in which there are 12 layers, 768 hidden states and 12 attention heads in the transformer. We truncate the BERT input to 256 tokens and fine-tune the BERT parameters during training.
We adopt the Adam~\citep{Adam} optimizer with the initial learning rate of 0.001. We apply the dropout~\citep{Dropout} regularization with the dropout probability of $0.2$ to reduce over-fitting. The batch size is 32. The training epoch is 60 with early stop. The weight of $L_2$ regularization is 0.015. The learning rate begins to decay after 10 epoch. The overlap rate of data samples is 50\%, and the number of selected news in each group is 3. 
When splitting the dataset, we guarantee that the samples in train set are previous to samples in valid set and test set to avoid the possible information leakage.
We tune the hyper-parameters on the development set and test model on the test set. 
The forex prediction is conducted as a binary classification task (up or down). The evaluation metrics are macro-F1 and Matthews Correlation Coefficient (\textbf{MCC}). MCC is often reported in stock movement forecast~\citep{stock_acl19_vae,stock_ding_event3} because it can overcome the data imbalance issue. 

\section{Results and Analysis}
\subsection{Comparison with Baselines}
\label{section-baseline}
Here, we introduce the baselines in this work. Since there are few existing works, we modify two advanced models from stock prediction field which adopt multi-news as input for this task. Besides, we design some ablation variations of the proposed model to check the effects of different modules. The baselines are shown below:
\begin{itemize}
    \item \textbf{NoNews:} This method considers the forex trade data only and use a 3-layer perceptron (the setting is same as full model) to encode the trade data and make prediction. This is a baseline to check the improvement by adding text information.
    \item \textbf{SVM:} This method chooses the support vector machine to predict the result based on the feature vectors extracted by the method introduced in~\citep{text_wsd_sa}.
    \item \textbf{HAN:} This method is proposed in~\citep{stock_hann} for stock movement prediction. It includes a hybrid attention mechanism and Gated Recurrent Unit to combine multi-day's stock news to predict movement. We use every 5 minutes instead of each day as time unit for this method and the StockNet method because there is too much news for forex trading and the experiments show that the latest news has the most influence.
    \item \textbf{StockNet:} This method is proposed in~\citep{stock_acl19_vae}. It treats the prediction task as a generation task and designs a modified variational auto encoder to process multi-days' tweets to predict stock movement. 
    \item \textbf{NoGroup:} This method does not group news and select key news directly from all news.
    \item \textbf{NoConnect:} This method does not connect news in the same group. Instead, it gets the representation for each news independently using BERT. This method groups news by category. 
    \item \textbf{LSTM+Attention:}  This method uses the bi-directional LSTM and self-attention to replace the BERT as text encoder. The number of LSTM hidden states is 256, and the hidden-layer is 3. This method groups news by category. 

\end{itemize}

\begin{table*}[t]
\centering
\scalebox{0.95}{
\begin{tabular}{l|cc|cc|cc|cc}
        & \multicolumn{2}{c|}{USD-EUR}   & \multicolumn{2}{c|}{USD-JPY}   & \multicolumn{2}{c|}{USD-RMB}  & \multicolumn{2}{c}{USD-GBP}    \\ \hline
\multicolumn{1}{c|}{Method}          & F1            & MCC            & F1            & MCC            & F1            & MCC           & F1            & MCC            \\ \hline
NoNews        & 63.0            & 0.266        & 64.8          & 0.295        & 65.4          & 0.304             & 64.7          & 0.301          \\
SVM         & 64.8          & 0.297         & 65.7          & 0.314        & 66.2          & 0.324           & 65.1          & 0.310           \\
HAN          & 65.2          & 0.305        & 67.0            & 0.341        & 66.7          & 0.334            & 66.9          & 0.346          \\
StockNet       & 65.4         &   0.309           & 66.8           &      0.336       & 67.2        &0.343    & 66.5        &   0.339      \\
NoGroup        & 66.7          & 0.335        & 67.5          & 0.350       & 68.0            & 0.361              & 68.3          & 0.375          \\
NoConnect    & 68.8     & 0.377            &70.9       &   0.418    & 69.6        & 0.392        & 68.7         &   0.383      \\
LSTM+Attention     & 69.8          & 0.397          & 71.2         & 0.422      & 71.8          & 0.434          & 69.7          & 0.403         \\ \hline
BHAM-Time      & 70.7          & 0.414        & 70.5          & 0.409       & 71.4          & 0.426           & 69.2         & 0.392          \\
BHAM-Topic      & 71.8          & 0.436        & 72.6          & 0.451       & 72.3          & 0.445         & 71.3          & 0.435          \\
BHAM-Category & \textbf{72.5} & \textbf{0.450}   & \textbf{73.4} & \textbf{0.466}  & \textbf{73.5} & \textbf{0.468} & \textbf{71.6} & \textbf{0.441}
\end{tabular}}
\vspace{-0.1in}
\caption{Results of baselines and proposed methods on the test set (input time window is 40 minutes, and prediction delay is 5 minutes, we observe similar result in other time settings). All the experiment results have proven significant with $p<0.05$ by student t-test.}
\label{result_baseline}
\vspace{-0.2in}
\end{table*}

As shown in Table~\ref{result_baseline}, all the three proposed methods perform well, and both BHAM-Topic and BHAM-Category methods outperform all the baselines. 
The BHAM-Category performs best among these methods, which shows that the semantic information of finance news is mostly aggregated by category. 
All the methods get improved after introducing the text information, which proves the related finance news is helpful for the prediction.
The performance of NoGroup method decreases by a large margin compared to BHAM-Category, which demonstrates that the hierarchical structure works well. Without hierarchical structure, selecting essential news directly from all news has more noise and requires the model to have a stronger fitting ability for a longer paragraph. 
After removing the news connection, the performance of NoConnect method drops sharply compared to BHAM-Category. Accessing the news representation from the connected paragraph helps the news representation realize the context information in the group.
The LSTM+Attention method performs worse than the BERT-based method, which proves that BERT has stronger power of sentence encoding. 
The two methods borrowed from stock movement prediction are designed to consider all news's information, but the forex related news is redundant, which can explain the poor performance of these two methods.

\subsection{Effect of Time Parameters}

\begin{figure}[t]
\centering
\includegraphics[scale=0.52]{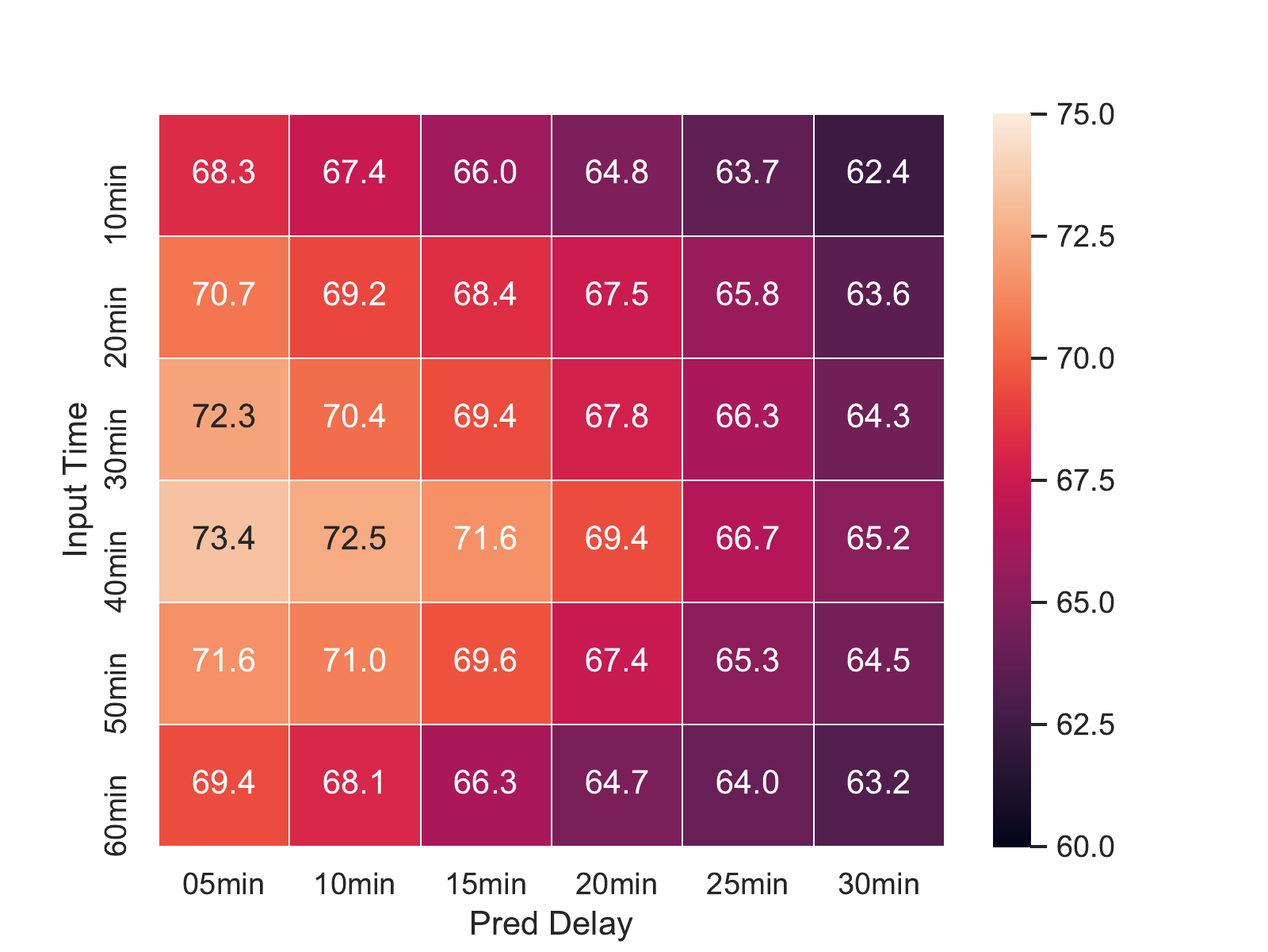}
\vspace{-0.1in}
\caption{The BHAM-Category model's performances (macro-F1\%) on USD-JPY pair under different conditions of input time and prediction delay. The dark colour means low performance and light colour means high performance.}
\label{result_matrix}
\vspace{-0.2in}
\end{figure}

In this section, we analyze the influence of two crucial time parameters on model performance, which are input time and prediction delay. We choose the input time $\in \left \{10, 20, 30, 40, 50, 60\right\}$ (minutes), the prediction delay $\in \left \{5,10,15,20,25,30\right\}$ (minutes) and experiment all combinations.
We take the USD-JPY for example to analyze the time effect of forex trading, and we observe similar results in other currency pairs.
The Figure~\ref{result_matrix} shows BHAM-Category model's performances (macro-F1\%) on USD-JPY pair under different combinations of input time and prediction delay. We can observe that with the increase of input time from 10 minutes to 40 minutes, the model performance improves too. However, when we increase the input time continuously, the model performance begins to decrease. Too less text is not enough to support the prediction, but too many texts may bring much noise. The ideal input time is around 40 minutes. Besides, at all input time conditions, the model's performances decline with the increase of prediction delay because events happened in the prediction delay time may also influence the forex movement. We can also conclude that forex movement pays more attention to the latest news because when masking the latest news input (such as USD-JPY(40-05) and USD-JPY(30-15), the latter one can be seen as the former one masking the lastest 10 minutes input), the model performance declines obviously at almost all conditions.

\begin{figure*}[t]
\centering
\includegraphics[scale=0.41]{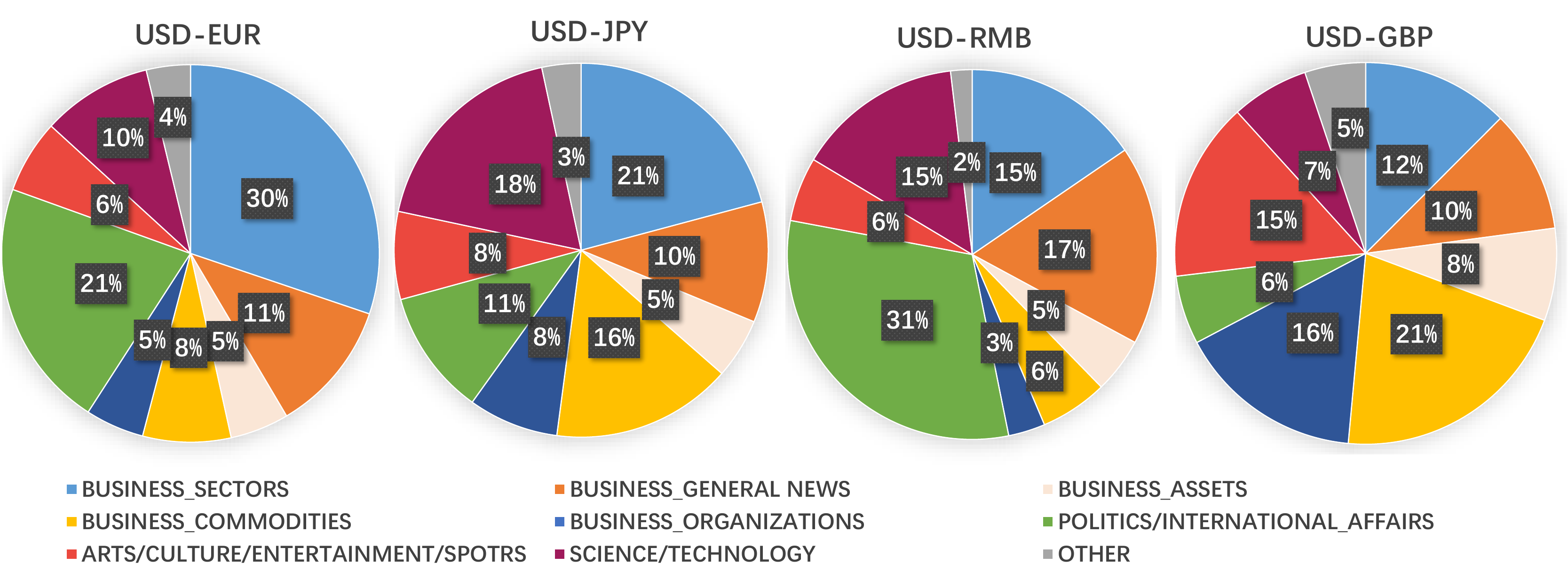}
\vspace{-0.1in}
\caption{The attention distributions over categories for different currency pairs.}
\label{category}
\vspace{-0.15in}
\end{figure*}

\begin{figure*}[t]
\centering
\includegraphics[scale=0.43]{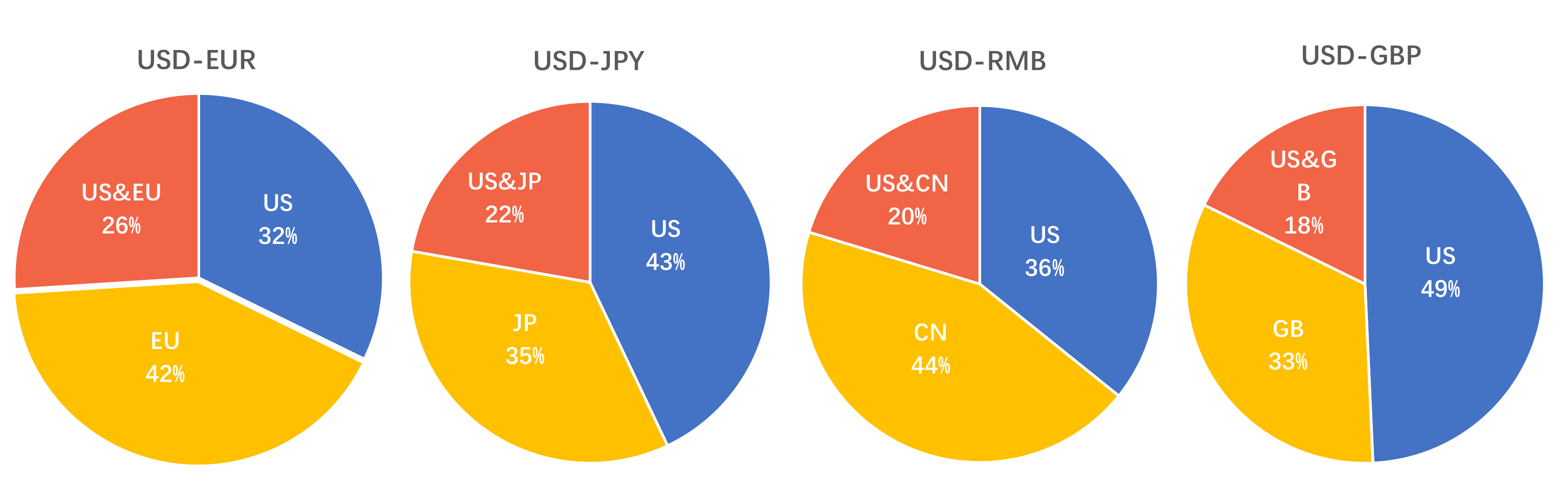}
\vspace{-0.1in}
\caption{The attention distributions over regions for different currency pairs.}

\label{region}
\vspace{-0.2in}
\end{figure*}

\subsection{Influence of News Attributes}
In this section, we analyze the influence of finance news's attributes (category and region) on prediction results and summarize the influence patterns for different currency pairs. We conduct the experiments based on BHAM-Category.
\subsubsection{Effect of News Category}
The forex trading data's attention weights over news categories are calculated by Equation~\ref{co-attention}. We sum up all the attention weights of test samples and calculate the proportions each category contributes. 
As shown in Figure~\ref{category}, we display the influence patterns of news category for different currency pairs. We observe that there are obvious differences among currency pairs. USD-EUR trading pays more attention to the Business Sectors and Politics/International Affairs news. USD-JPY trading is mostly influenced by Business Sectors and Science/Technology news. Politics/International Affairs news has the most significant impact on USD-RMB trading and Business Commodities news effects USD-GBP trading most. The summarized influence patterns can serve as decision-making reference for forex traders when facing news from various categories.

\subsubsection{Effect of News Region}
The trading data's attention weight for selected news $att_{ij}$ is calculated by the following formula:
\begin{flalign}
&att_{ij} = att_{i} * s^i_{j}  
\end{flalign}
Where $att_{i}$ is the trade data's attention on the i-th category in Equation~\ref{co-attention} and $s^i_{j}$ in Equation~\ref{intra-weight} is the weight of selected news in group. We sum up all the selected news's attention according to their regions and access the region influence weight. The results are shown in Figure~\ref{region}. For each currency pair, the news are divided into three classes: news related to region A only, news related to region B only and news related to both region A and B. And we observe that the news related to both region A and B has the least influence on all currency pairs. News related to the US has the largest influence weight on USD-JPY and USD-GBP trading. Yet news related to China/Europe has a larger influence weight than news related to US in USD-RMB/USD-EUP trading. We can intuitively observe the influence weights of different regions for forex trading, which is helpful for the analysis and forecast of forex movement.


\begin{table}[]
\centering
\scalebox{0.8}{
\begin{tabular}{c|cccc}
  & \textbf{USD-EUR} & \textbf{USD-JPY}  & \textbf{USD-RMB} & \textbf{USD-GBP} \\ \hline
\textbf{1}   & 67.6              & 68.8      & 69.3                  & 67.3             \\
\textbf{2}     & 71.4          & 73.1         & 72.2                 & 70.8             \\
\textbf{3}    & $\bm{72.5}$   &  $\bm{73.4}$        &  $\bm{73.5}$                  & $\bm{71.6}$            \\
\textbf{4}     & 72.2              & 72.8          & 73.1            & 70.7             \\
\textbf{5}     & 70.8        & 70.3        & 71.9                    & 68.4             \\
\bm{$\infty$ }   & 64.1   & 64.5   & 65.7     & 63.6  
\end{tabular}}

\caption{Impact of selection number in each group in BHAM-Category. $\infty$ means keeping all news. The results have proven statistic significant.}
\label{result_sele}
\vspace{-0.2in}
\end{table}

\subsection{Impact of Selection Number}
The selection number in each group is an essential hyper-parameter to control the amount of extracted information.  
As shown in Table~\ref{result_sele}, the BHAM-Category performs best when the selection number is 3 in all currency pairs. When the selection number is small (1,2), the model is too strict so that some crucial information will be missed. When the selection number is large (4,5), some less influential news will be selected and interfere model's decision. When we keep all news in the group, the model's performance declines by a large margin. This experiment demonstrates that the selection mechanism plays an important role in the proposed model. 

\section{Conclusion}
In this work, we propose a BERT-based Hierarchical Aggregation Model to summarize a large amount of finance news for forex movement prediction. Experiments show that our model outperforms all the baselines by a large margin, which proves the effectiveness of the proposed framework. We design three grouping news methods: time, topic and category and experiments show that the category-based method performs best, which shows that the semantic information of forex related  news is mostly aggregated by category. 
Experiments about time effect prove that the proper input time is about 40 minutes and the prediction accuracy declines with the increase of prediction delay. Besides, we analyze the influence of news attributes on forex trading and observe some interesting conclusions: Business Sectors news has the most influence on USD-EUR trading and Politics/International Affairs news effects USD-RMB trading most. Besides, both USD-JPY trading and USD-GBP trading pay most attention to news from US. 
All these influence patterns can help forex traders handle different news more wisely and make better decisions.

To our knowledge, this is the first work to utilize the advanced NLP pre-train
technology in the enormous forex market and the results show the potential of this research area.
Promising future studies may include designing more suitable grouping methods or combining news grouping and market predicting in an end2end model. 
\section{Acknowledgement}
This work is supported by a Research Grant from Mizuho Securities Co., Ltd. Mizuho Securities also provide experiment data and valuable domain experts suggestions. 

\bibliography{emnlp-ijcnlp-2019}
\bibliographystyle{acl_natbib}

\appendix

\end{document}